\documentclass[lettersize,journal]{IEEEtran}
\usepackage{amsmath,amsfonts}
\usepackage{algorithmic}
\usepackage{algorithm}
\usepackage{array}
\usepackage[caption=false,font=normalsize,labelfont=sf,textfont=sf]{subfig}
\usepackage{textcomp}
\usepackage{stfloats}
\usepackage{url}
\usepackage{verbatim}
\usepackage{graphicx}
\usepackage{cite}
\usepackage{microtype}
\usepackage{graphicx}
\usepackage{amsmath}
\usepackage{booktabs}
\usepackage{amssymb}
\usepackage{amsfonts}
\usepackage{mathtools}
\usepackage[mathscr]{euscript}
\usepackage{xcolor}
\usepackage{color}
\begin{document}

\title{\textit{CET2}: Modelling Topic Transitions for Coherent and Engaging Knowledge-Grounded Conversations}
\author{Lin Xu, Qixian Zhou, Jinlan Fu$^{*}$\thanks{\ \ * Corresponding author}, See-Kiong Ng

\thanks{Lin Xu, and See-Kiong Ng are with the Department of Computer Science, School of Computing, National University of Singapore and together with Jinlan Fu, are with Institute of Data Science, National University of Singapore, 117602. Qixian Zhou is with Bytedance, Shenzhen, China. (E-mail: cathyxl2016@gmail.com, qixianzhou@gmail.com, jinlanjonna@gmail.com, and seekiong@nus.edu.sg)}
}
% \markboth{IEEE/ACM Transactions on Audio, Speech, and Language Processing,~Vol.~31, No.~8, September~2023}%


\maketitle

\begin{abstract}
Knowledge-grounded dialogue systems aim to generate coherent and engaging responses based on the dialogue contexts and selected external knowledge. Previous knowledge selection methods tend to rely too heavily on the dialogue contexts or over-emphasize the new information in the selected knowledge, resulting in the selection of repetitious or incongruous knowledge and further generating repetitive or incoherent responses, as the generation of the response depends on the chosen knowledge. To address these shortcomings, we introduce a Coherent and Engaging Topic Transition (CET2) framework to model topic transitions for selecting knowledge that is coherent to the context of the conversations while providing adequate knowledge diversity for topic development. Our CET2 framework considers multiple factors for knowledge selection, including valid transition logic from dialogue contexts to the following topics and systematic comparisons between available knowledge candidates. Extensive experiments on two public benchmarks demonstrate the superiority and the better generalization ability of CET2 on knowledge selection. This is due to our well-designed transition features and comparative knowledge selection strategy, which are more transferable to conversations about unseen topics. Analysis of fine-grained knowledge selection accuracy also shows that CET2 can better balance topic entailment (contextual coherence) and development (knowledge diversity) in dialogue than existing approaches.
\end{abstract}


\section{Introduction}
% \IEEEPARstart{T}{his} file is intended to serve as a ``sample article file''
% \jlfu{Cite some papers that are published or released in 2022.}
\begin{figure*}[t]
\centering
  \includegraphics[width=0.7\textwidth]{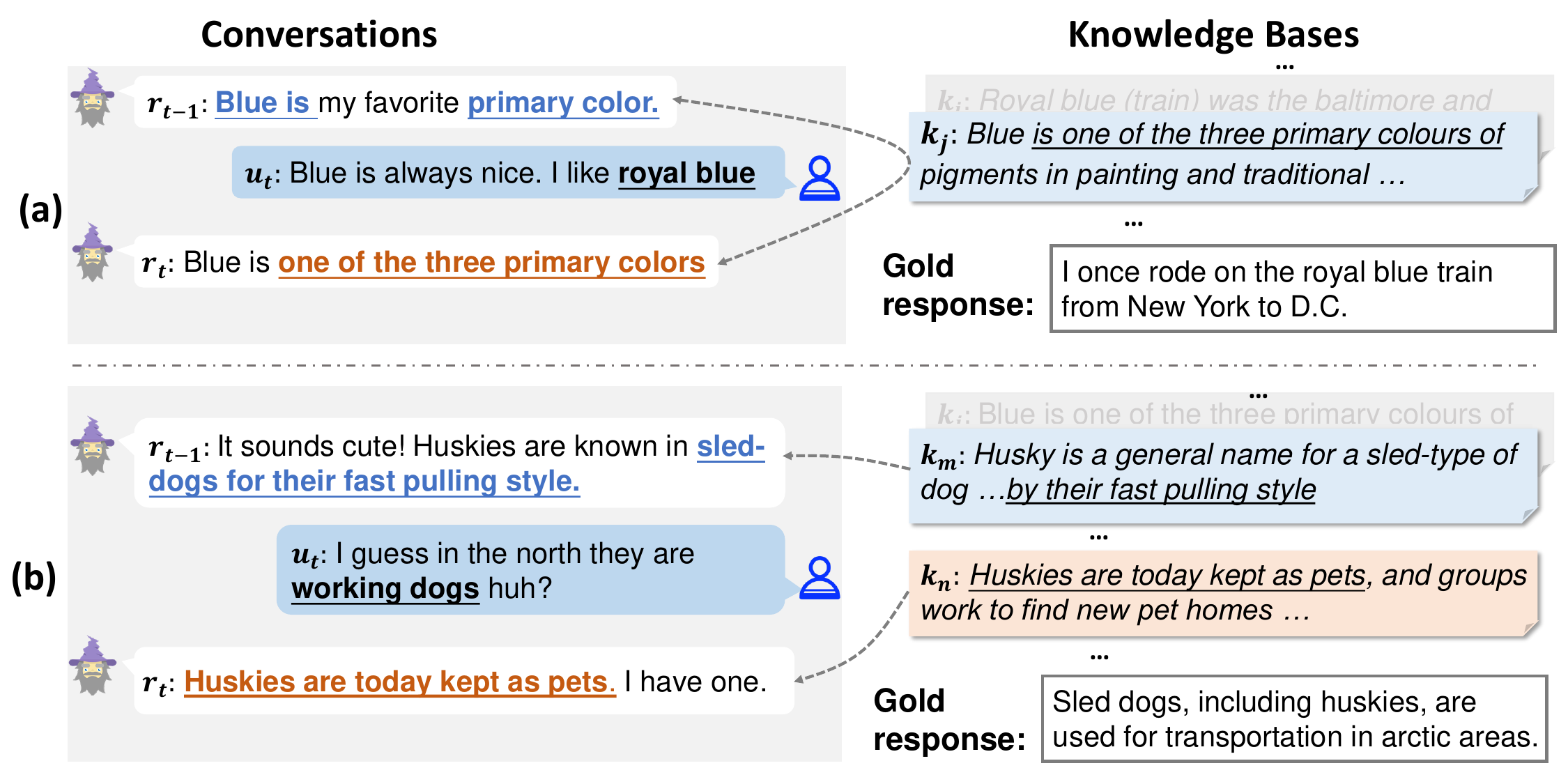}
\caption{Two conversation examples of knowledge selections lacking in diversity (a) or coherence (b) and leading to non-engaging or incoherent conversations. $r_{t-1}$ and $u_t$ are utterances from dialogue history. Knowledge bases are the knowledge candidates to be chose from. $k_j$ was chosen for $r_{t-1}$ and $r_t$ in (a), leading to repetitive responses. $k_m$ and $k_n$ are chosen for $r_{t-1}$ and $r_t$ in (b), 
}
\label{fig:intuition}
\end{figure*}
\IEEEPARstart{A}{} key challenge for open-domain dialogue agents is to generate informative response~\cite{ghazvininejad2018knowledge, zhou2020design} that satisfy humans' need for information in communication. To explore dialogue agents' ability of applying external knowledge,  the task knowledge-grounded conversations~\cite{dinan2019wizard} are proposed, where conversation task is divided into two steps, knowledge selection and response generation. The former selects the next knowledge to be used from a pool of knowledge candidates based on the dialogue context, while the latter then generates a natural language response based the selected knowledge.

We focus on knowledge selection, which is also known as the topic transition modelling problem. 
The skillful usage of knowledge in dialogue systems is particularly important for generating engaging knowledge-grounded conversations. The knowledge used to generate a response should be related to the context of the ongoing conversation (i.e. \textbf{coherent}) as well as sufficiently diversified to engage the user's interests (i.e. \textbf{engaging}). Merely injecting new knowledge into a generated response does not necessarily improve the quality of a conversation, while repetitive use of similar knowledge may disengage the user prematurely.

Most of the existing methods~\cite{lian2019learning,kim2020sequential,zheng2020approximation,zhao2020knowledge} directly rely on the dialogue context to select the next knowledge. 
Without appropriate topic transition modeling, spurious correlations between dialogue context and knowledge may result in knowledge selection that lacks in diversity or coherence and in the mean time hard to be extended to unseen scenarios. As shown in the top example in Figure~\ref{fig:intuition} (a), the method~\cite{zhao2020knowledge} selects the knowledge $\mathbf{k}_j$ (\emph{``blue is one of the three primary color...''}), which repeats the response $\mathbf{r}_{t-1}$ (\emph{``blue is my favorite primary color''}) in the conversation's first turn, thus introducing no new knowledge that interests the user. To ensure the introduction of new knowledge into the conversation, a study~\cite{zheng2020difference} considers the differences between the knowledge used in two consecutive turns. However, focusing only on the knowledge differences for knowledge selection could sacrifice the dialogue coherence. As shown in the bottom example in Figure~\ref{fig:intuition} (b), the predicted (i.e. selected) knowledge $\mathbf{k}_m$ about ``\emph{huskies as pet}'' shifts too much away from the ongoing dialogue context $\mathbf{r}_{t-1}$ (\emph{``Huskies are known for...fast pulling style''})  and $\mathbf{u}_t$ (``\emph{working dogs}''), as there were no sign of topic change by the user who was still asking whether huskies are working dogs. 

\begin{figure}[t]
\centering
  \includegraphics[width=0.48\textwidth]{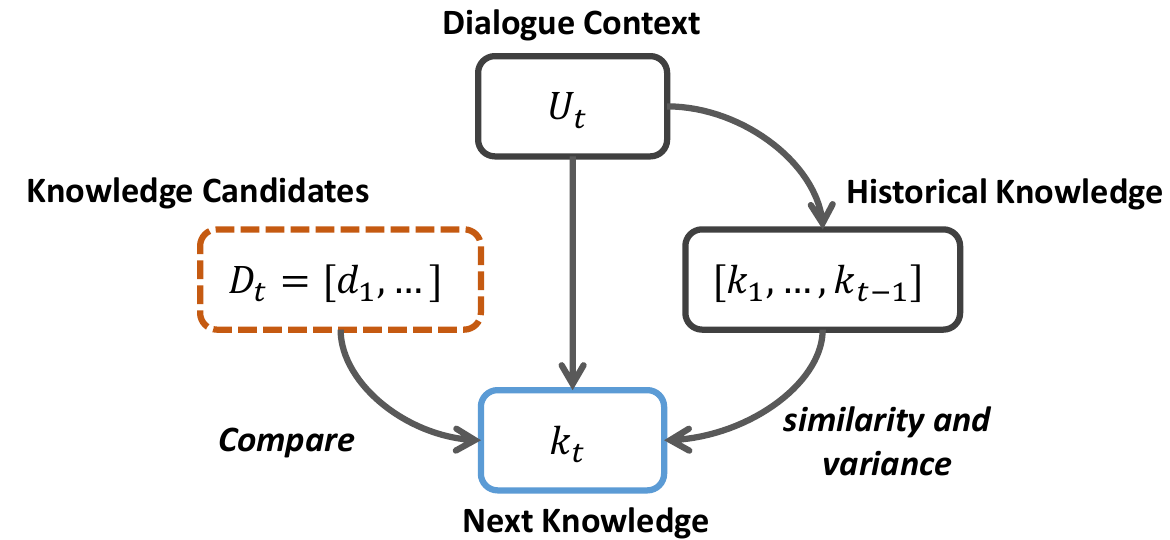}
\caption{Mechanisms for selecting knowledge in response considering both dialogue history and knowledge candidates.}
\label{fig:select_mecha}
\end{figure}

Taking into consideration the above topic transition problems of existing methods, we propose a novel knowledge selection framework called CET2 (\textbf{C}oherent and \textbf{E}ngaging \textbf{T}opic \textbf{T}ransition) to comprehensively decide the next knowledge by including all relevant factors in knowledge grounded conversations. We posit that the selection of appropriate knowledge to generate the next response depends not only on the apposite sequential (or "vertical") transitions within the dialogue contexts but also on the "horizontal" semantic comparisons among all knowledge candidates, as shown in Figure~\ref{fig:select_mecha}, where three factors, \textbf{dialogue context, historical knowledge, and knowledge candidates} simultaneously decide the next knowledge.  Dialogue history and historical knowledge provides the context, based on which the model should select the next knowledge. As for the existing knowledge candidates, intuitively, when people find it hard to think what to talk about next, comparatively selecting a good one from a set of knowledge candidates could also solve the situation.

For CET2, we mainly develop a comparative knowledge selection method, which comprehensively selects knowledge by combining our well-designed topic transition features that ensure both topic coherence and diversity, and the relational feature of all knowledge candidates. Additionally, optimized in a variance-aware training strategy, CET2 captures suitable knowledge variance in consecutive turns and thus consequently contributes to both coherent and engaging responses.

In summary, our contributions are threefold. 
\begin{itemize}
    \item  We propose a novel CET2 knowledge selection framework that models appropriate topic transitions by simultaneously the topic coherence and variance, as well as comprehensive comparisons of all the knowledge candidates.
    \item  We demonstrate the superiority and better generalizability of CET2 in knowledge selection, outperforming the previous SOTA by 1.6\% and 4.7\% in seen and unseen scenarios respectively.
    \item We analyze the topic adhesion and diversity in knowledge selection by our proposed automatic metrics and conducting a human evaluation of the generated conversations. The human-annotated results can further be used as a source for learning dialogue coherence/diversity.
\end{itemize}

\section{Related Work}
\subsection{Knowledge-Grounded Conversation.}
Knowledge is a key part of any kind of conversation, so it is an essential ability for conversation agent to combine external knowledge~\cite{moon-etal-2019-opendialkg, xu-etal-2020-conversational} implicitly~\cite{ghazvininejad2018knowledge} or explicitly~\cite{dinan2019wizard}. To explore the knowledge flows in knowledge-rounded conversations, ~\cite{dinan2019wizard} first presents a benchmark where knowledge is explicitly labeled for each conversation turn, which makes possible the measurement of dialogue agents' topic modeling ability through calculating the knowledge selection accuracy. Under this task setting, many relevant works~\cite{lian2019learning, zheng2020approximation, zhao2020knowledge} propose neural networks to rank knowledge candidates with dialogue contexts. For example, \cite{lian2019learning} learns knowledge selection distribution by minimizing its distance with the posterior knowledge distribution conditioned on responses. Similarly, \cite{zheng2020approximation} also committed to make close knowledge selected with dialogue contexts and that of response-retrieved on both word and sentence levels. Some other recent works~\cite{li-etal-2022-enhancing-knowledge, xu-etal-2022-corefdiffs} further refer to extra document structures information to do knowledge selection to boost knowledge selection accuracy but no relevant research on balancing topic coherence and development in this task.

With the selected knowledge and dialogue context as input, these models then adopt common language decoders~\cite{vaswani2017attention, radford2019language} to generate responses. As there are numerous studies~\cite{zhao2020knowledge, lin2020generating, rashkin-etal-2021-increasing} addressing response generation with knowledge, we focus on the knowledge selection task in this paper and adopt GPT-2~\cite{radford2019language} as our response generator. 

\subsection{Topic Transition Modeling.}
Knowledge selection can be treated as a topic transition modeling task, which is from dialogue contexts, and historical knowledge to the next knowledge to be used in responses. ~\cite{kim2020sequential} proposes to implicitly historical knowledge sequences by latent variables. ~\cite{zheng2020difference} focuses on the difference between historical knowledge and the next one. ~\cite{meng2020dukenet} designs a knowledge tracker and shifter to model knowledge interactions between turns and adopts dual learning to optimize the whole model. ~\cite{zhan2021augmenting} abstracts topic labels for the knowledge to reduce sequential transition noises. However, the challenge of maintaining well-balanced coherence and knowledge diversity in knowledge-grounded conversations, which was also highlighted in traditional dialogue system studies~\cite{li-etal-2016-deep, li-etal-2017-end}, was not addressed by these methods. In this paper, we propose an effective and robust knowledge selection method to generate dialogue responses that are coherent to the dialogue context while introducing suitably new knowledge to keep users engaged in the conversations.  

\section{Method}
\subsection{Task Formulation}
In a knowledge-grounded dialogue between a user and an agent (the chatbot), the agent's next response $r_t$ is to be generated given the current dialogue context $U_t=\{u_{t-l},r_{t-l},...,r_{t-1}, u_t\}$ and a pool of knowledge candidates $D_t=\{d_{t,j}\}^M_{j=1}$, where $t-l,...,t$ are the turn indexes and $l$ is the length of the dialogue context in terms of number of turns. $u_\ast$ and $r_\ast $ are utterances from the user and the agent, respectively. $d_{t,\ast}$ are text-based knowledge sentences to be chosen from in the current turn. For simplicity, let us omit the subscript $t$ in subsequent discussions. The two steps of knowledge-grounded conversation, knowledge selection, and response generation,  are thus formulated as $P(\hat{d}|U, D)$ and $P(r|U, \hat{d})$ respectively, where $\hat{d}$ is the selected knowledge by the selection model. For a better description of training objectives, we denote the ground-truth knowledge as $\Tilde{d}$.

\begin{figure*}[t]
\centering
  \includegraphics[width=0.88\linewidth]{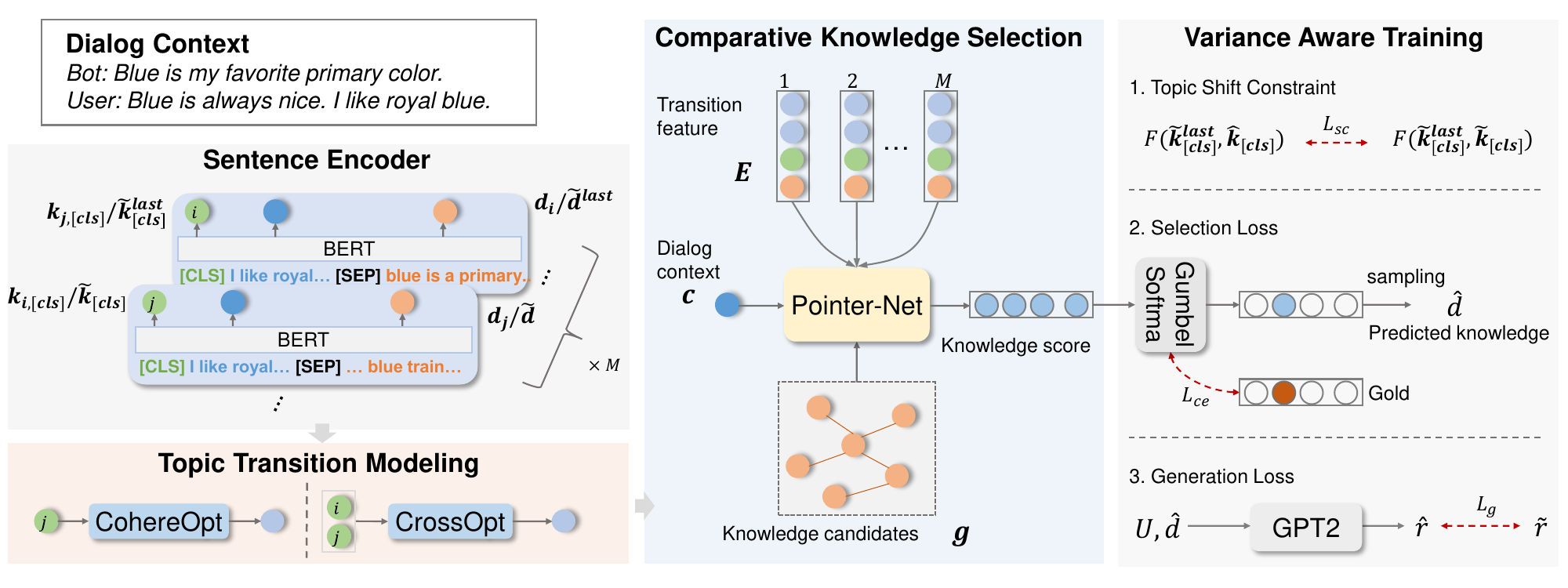}
\caption{The architecture of CET2. The Sentence Encoder outputs representations for the dialogue context and knowledge candidates. Two special candidates are shown: the $i$-th knowledge $d_i$ (also denoted $\Tilde{d} $), is the gold knowledge for the current turn and the $j$-th knowledge $d_j$ (denoted $\Tilde{d}^{last}$) is the gold knowledge of last turn. The Topic Transition Modeling acquires transition features. The Comparative Knowledge Selection (in blue) outputs the selected knowledge. The Variance Aware Training depicts our training strategy of controlling the knowledge variance between turns with Topic Shift Constraints.}
\label{fig:framework}
\end{figure*}

\subsection{CET2 Framework}
Figure~\ref{fig:framework} shows the overall architecture of our CET2 knowledge selection framework. It mainly consists of four parts, namely Sentence Encoder for dialogue history and knowledge encoding, Topic Transition Modeling to capture transition features, Comparative Knowledge Selection comprehensively performing knowledge selection with three factors considered, and Variance-aware Training denoting the training objectives.

\subsubsection{Sentence Encoder}
Similar to~\cite{zhao2020knowledge}, we adopt BERT~\cite{devlin2018bert} to obtain the embeddings for the dialogue context and each knowledge candidate as follows. The utterances in the dialogue context $U$ are concatenated together into a long text $C=\text{[usr]}u_{t-l}\text{[agt]}r_{t-l}...r_{t-1}\text{[usr]}u_t$, where [usr], [agt] are used to divide utterance from the user or the chatbot. $C$ is then combined with each knowledge candidate separately to form $M$ paired inputs to BERT. We denote this set of paired inputs as $I$, which is defined as $I =\{\text{[CLS]} C\text{[SEP]}\} d_j \}_{j=1}^M$.

The paired inputs in $I$ are then fed into BERT to yield their representations, as shown in Figure~\ref{fig:framework}. For each paired input $I_j$, we obtain the hidden state for each token after BERT encoding. The hidden state of the special token $\text{[CLS]}$, denoted as $\mathbf{k_{j,[cls]}} \in \mathbb{R}^d$, is a \textbf{context-aware knowledge representation} of the $j$-th knowledge candidate. In fact, $\mathbf{k_{j,[cls]}}$ not only incorporates the information of the dialogue context $C$ and the knowledge $d_j$ but also embodies the semantic relations between them, such as entailment or other transitional relations.  Additionally, we extract the separate representations for each knowledge candidate by average pooling on their corresponding tokens, denoted as $\mathbf{k^j} \in \mathbb{R}^d$. Similarly, we obtain the representation of the dialogue context $C$, denoted as $\mathbf{c_j} \in \mathbb{R}^d$. The process for each paired input $I_j$ is formulated as:
\begin{equation*}
    \mathbf{k_{j,[cls]}},\mathbf{c_j},\mathbf{k_j} \in \mathbb{R}^d =\text{BERT}(I_j), j\in[1,..M]
\end{equation*}
We then aggregate all the $M$ dialogue context representations $\{\mathbf{c_j}\}_1^M$ by an attention operation to obtain the aggregated context representation $\mathbf{c} \in \mathbb{R}^d$ that fuses all knowledge candidates: 
\begin{equation*}
\begin{aligned}
    \mathbf{h_j}&=\tanh(W_{c}\mathbf{c_j}), &\\
    {\alpha}_j&=\frac{\exp(V_{c}\mathbf{h_j})}{\sum_{i=1}^{M}\exp(V_{c}\mathbf{h_i})},& \\ 
    \mathbf{c}&=\sum_{i=1}^{M}{{\alpha}_i}{\mathbf{h_i}},
\end{aligned}
\end{equation*}
\noindent where $W_c\in \mathbb{R}^{d\times d}$ and $V_c\in \mathbb{R}^{d}$ are trainable weights.

\subsubsection{Topic Transition Modeling}
We model the topic transition mechanism in knowledge-grounded dialogue systems to take into account both topic coherence and development given the dialogue history. Two transition features, topic entailment and topic development, are designed in CET2.
% We design two transition features of the knowledge candidates for predicting the next knowledge with the dialogue context.

%\noindent\textbf{Topic Entailment.} 
% \noindent\textbf{Topic Coherence.} 
\paragraph{Topic Coherence} 
We adopt the above-mentioned context-aware knowledge representation $\mathbf{k_{j,[cls]}}$ to obtain the topic coherence features for $j$-th knowledge candidate. We note that the output hidden state of the token $\text{[CLS]}$ in the BERT's last layer measures with the entailment correlations between the dialogue context and a knowledge candidate, thanks to BERT's Next Sentence Prediction pre-training scheme~\cite{devlin2018bert}.  We employ a single fully connected layer with $\tanh$ activation to get the \textbf{topic entailment feature}, denoted as:
\begin{equation*}
    \mathbf{v^{coh}_j} = \tanh(W_{coh}\mathbf{k_{j,[cls]}}) \in \mathbb{R}^{d_{coh}}
\end{equation*}
where $W_{coh}\in \mathbb{R}^{d_{coh}\times d}$ is trainable parameters. This is the CohereOpt, as shown in the Topic Transition Modelling module of Figure~\ref{fig:framework}.

% \noindent\textbf{Topic Development.}
\paragraph{Topic Development}
We adopt the inference feature computation method~\cite{chen2017enhanced} used in the Natural Language Inference task to capture the other transition-aware feature from the perspective topic development, the CrossOpt in Figure~\ref{fig:framework}. It models the high-order interactions between the two vectors by both element-wise difference and product. Specifically, we first obtain the context-aware knowledge representation $\mathbf{\Tilde{k}^{last}_{[cls]}}$ of the ground-truth knowledge in the previous turn from the Sentence Encoder. Then we compute the \textbf{topic development feature} $\mathbf{v^{cro}_j} \in \mathbb{R}^{d_{cro}}$ for a knowledge candidate by the cross operator followed a fully connected layer and $\tanh$ activation. The process is as follows:
\begin{equation*}
\begin{aligned}
\mathbf{v^{cro}_j}&=\sigma(W_{cro}([\mathbf{\Tilde{k}^{last}_{[cls]}}- \mathbf{k_{j,[cls]}};\mathbf{\Tilde{k}^{last}_{[cls]}}\odot\mathbf{k_{j,[cls]}}])) \in \mathbb{R}^{d_{cro}}\\
\end{aligned}
\end{equation*}
where $W_{cro} \in \mathbb{R}^{d_{cro}}$ is trainable parameters. We set $\mathbf{v^{cro}_j}$ to zero vector where there is no last knowledge, for example, at the first turn of a conversation.

\paragraph{Transition-aware Representation} 
The topic entailment feature, $\mathbf{v^{coh}_j}$, topic development feature $\mathbf{v^{cro}_j}$, context-aware knowledge representation $\mathbf{k_{j,[cls]}}$, and knowledge representation $\mathbf{k_j}$ together form the \textbf{transition features}, $\mathbf{E} \in \mathbb{R}^{M\times d_e}$ of the $M$ knowledge candidates, which is denoted as:
\begin{equation*}
\begin{aligned}
   &\mathbf{e_j} = [\mathbf{v^{coh}_j};\mathbf{v^{cro}_j};\mathbf{k_{j,[cls]}};\mathbf{k_j}] \\
    &\mathbf{E} = \{\mathbf{e_j}\}_{j=1}^M,
\end{aligned}
\end{equation*}
where $[;]$ means the concatenation operator along the last dimension of the tensor.

\subsubsection{Comparative Knowledge Selection}
Instead of individually matching the dialogue context with each knowledge candidate, we posit that it is important to compare all the knowledge candidates before selecting the next knowledge. PointerNet~\cite{vinyals2015pointer} is known for its ability to rank input in variable size, which is well-aligned without knowledge selection task needing a prior comprehension of knowledge candidates. In particular, we adapt PointerNet for our task by first encoding all knowledge and their relations with graph representation and then feeding them with the transition features into a single PointerNet step to obtain ranked knowledge selection scores.

We introduce a multi-head graph attention network (GAT)~\cite{velivckovic2018graph} to encode the associations of all the knowledge candidates within a graph structure instead of the sequence modeling in the vanilla PointerNet. The graph structure $\mathcal{G}$ is constructed based on the text-similarity (tf-idf) of the knowledge candidates. Each node is initialized by the knowledge representation $\mathbf{k_j}$. GAT The graph representation $\mathbf{g}$ for all of the knowledge candidates is obtained by average pooling, formulated as:
\begin{equation*}
    \mathbf{g}=\text{avgpool}(\text{FFN}(\text{GAT}([\mathbf{k_j}]_1^M, \mathcal{G}))) \in \mathbb{R}^d.
\end{equation*}
PointerNet is a sequence decoder, but we set the decoding length to 1 in this task to only choose one knowledge. In other words, with dialogue context $\mathbf{c}$ as query, transition-aware knowledge representations $\mathbf{E}$ as keys, and graph representation $\mathbf{g}$ as the encoder hidden state, the PointerNet ranks knowledge candidates with score $p(d_j|U,D)$ as follows:
\begin{equation*}
\begin{aligned}
    &\mathbf{g'}=\text{LSTM-Cell}(\mathbf{c}, \mathbf{g}),\\
    &\beta_j=\mathbf{v}^{\top}\tanh(W_e\mathbf{e_j}+W_g\mathbf{g'}+\mathbf{b}),\\
    &p(d_j|U,D)=\frac{\exp(\beta_j)}{\sum_{i=1}^{M}\exp(\beta_i)},
\end{aligned}
\end{equation*}
where $W_e \in\mathbb{R}^{d\times{d_e}}$, $ W_g\in\mathbb{R}^{d\times{d}}$ and $\mathbf{v}, \mathbf{b}\in\mathbb{R}^{d}$ are trainable weights. 

%We select the knowledge with the maximum score $p(d_j|U,D)$ through Gumbel-Softmax~\cite{jang2016categorical} to ensure the differentiability.

\subsubsection{Variance-aware Training} 
To ensure suitable topic development, we introduce how to train the CET2 model with a variance-aware training strategy.

\paragraph{Topic Shifting Constraint.} 
We devise a Topic Shifting Constraint to explicitly control the variance between knowledge in consecutive turns. The constraint is an auxiliary loss in training phase given the context-aware knowledge representations of the ground-truth knowledge of the last turn, $\mathbf{\Tilde{k}^{last}_{[cls]}}$, the ground-truth knowledge and the predicted knowledge of the current turn, $\mathbf{\Tilde{k}_{[cls]}}$ and $\mathbf{\hat{k}_{[cls]}}$. To ensure the model's differentiability after adding the proposed topic shifting constraint, we select the knowledge with the highest score through Gumbel-Softmax~\cite{jang2016categorical}. In particular, we compute the information variance of two tuples, $\langle{\mathbf{\Tilde{k}^{last}_{[cls]}},\mathbf{\hat{k}_{[cls]}}}\rangle$ and $\langle{\mathbf{\Tilde{k}^{last}_{[cls]}},\mathbf{\Tilde{k}_{[cls]}}}\rangle$. The former measures the variance between the current selected knowledge and the previous ground-truth knowledge, while the latter computes the variance between the current and the previous ground-truth knowledge. Our variance-aware training strategy requires that these two distributions to be close to each other in order to sustain the exact variance between the historical knowledge and current knowledge. We adopt the Kullback-Leibler divergence to narrow down the difference of these two distributions, denoted as loss $L_{sc}$: 
\begin{equation*}
L_{sc} = D_{KL}(F(\mathbf{\Tilde{k}^{last}_{[cls]}},\mathbf{\hat{k}_{[cls]}}) {\Vert} F(\mathbf{\Tilde{k}^{last}_{[cls]}},\mathbf{\Tilde{k}_{[cls]}}))
\end{equation*}
We define the variance measure function $F$ as:
\begin{equation*}
    F(\mathbf{u},\mathbf{v})=\log\text{\_softmax}([(\mathbf{u}-\mathbf{v})^2;\mathbf{u}{\odot}\mathbf{v}])
\end{equation*}
where $\text{softmax}$ and $\log$ function ensures the two variances are two distributions with a sum equal to 1 and $\mathbf{u}$ and $\mathbf{v}$ are two vectors of the same dimension.

\paragraph{Knowledge Selection}

In a knowledge-grounded conversation data set, the turns of all the conversations form the data samples $\mathcal{D}={\{(U_i, D_i, r_i)\}}_1^N$ where $U_i$, $D_i={\{d_j^i\}}_{j=1}^M$ and $r_i$ are the dialogue context, knowledge candidates and next response respectively. $N$ is the total number of samples. The knowledge selector model is trained by minimizing the loss function $L_{cls}$ on $\mathcal{D}$ as follows:
\begin{equation*}
\begin{aligned}
    &L_{cls} = L_{ce} + \lambda L_{sc}\\
    &L_{ce} = -\frac{1}{N}\sum_{i=1}^{N}\Tilde{\mathbf{y}}_i\log(p(d_i|U_i,D_i))
\end{aligned}
\end{equation*}
where $\hat{\mathbf{y}}_i$ denotes the one-hot vector indicating the ground-truth knowledge for data sample $i$. $p(d_i|U_i,D_i)$ denotes the probability distribution over the candidate knowledge $D_i$. $L_{ce}$ is a standard cross-entropy loss function for knowledge selection and $\lambda$ is the coefficient that makes a balance between the two objective functions.

\paragraph{Response generation}
We generate response $r_t$ by fine-tuning GPT-2~\cite{radford2019language} model with the concatenated dialogue context $U_t$ and selected knowledge $\hat{d}_t$ as input. The GPT-2 model generates a distribution over the vocabulary $\mathcal{V}$ at each decoding position, which is optimized with the cross-entropy loss.
\begin{equation*}
\begin{aligned}
    &p(r_\tau^i|U_i,\hat{d}_i,r_{<\tau}^i) = \text{GPT-2}(\hat{d}_i, U_i,r_{<\tau}^i)\\
    &L_{g} = -\frac{1}{N}\frac{1}{|r_i|}\sum_{i=1}^{N}\sum_{\tau=1}^{|r_i|}\Tilde{\mathbf{y}}_\tau^i\log p(r_\tau^i|U_i,\hat{d}_i,r_{<\tau}^i)
\end{aligned}
\end{equation*}
where $\Tilde{\mathbf{y}}_\tau^{i} \in \mathbb{R}^{|\mathcal{V}|}$ is the one-hot vector indicating the ground-truth word at position $\tau$ of response. $p(r_\tau^i|U_i,\hat{d}_i,r_{<\tau}^i)$ is the probability distribution over the vocabulary at position $\tau$.

\section{Experiments}
\subsection{Datasets}
We evaluated our model on two commonly used public benchmark datasets for the knowledge-grounded dialogue system, Wizard of Wikipedia (WoW)~\cite{dinan2019wizard} and Holl-E~\cite{moghe2018towards}. WoW consists of 18430/1948/965/968 dialogs for train/valid/test\_seen/test\_unseen split. Each dialogue was constructed in a wizard-apprentice style, with the wizard (chatbot) trying to inform an apprentice (person) about a Wikipedia topic. Holl-E contains deeper conversations about Wikipedia movies. Following~\cite{kim2020sequential}, we adopted the 7211/930/913 splits for train/valid/test for Holl-E. The average dialogue turns, and candidate knowledge number per turn for WoW and Holl-E are 4, 5, and 61, 60, respectively.

\subsection{Evaluation Metrics}
Although we have focused on the knowledge selection task in this work,  we evaluate both the two sub-tasks of knowledge selection and response generation to compare with previous work. We use accuracy (\textbf{ACC}) to measure the performance of knowledge selection and further design two metrics Adhesion (\textbf{Adh.}) and Diversity (\textbf{Div.}) to evaluate the adherence and diversity of the knowledge. Specifically, we split all samples into \textbf{topic-adhesive} or \textbf{topic-changing} groups based on whether a data sample uses the same knowledge with its most recent historical turn or not. The two metrics are defined as the \textbf{ratio} of correct knowledge prediction within the above-mentioned two groups of samples, respectively, defined as:
\begin{equation*}
\begin{aligned}
    &Adh. = \frac{\text{\# right selection in topic-adhesive group}}{\text{\# topic-adhesive group}} \\
    &Div. = \frac{\text{\# right selection in topic-changing group}}{\text{\# topic-changing group}}
\end{aligned}
\end{equation*}

For the response generation, we use uni-gram F1(\textbf{uF1}), BLEU-1 \textbf{(B-1)}, BLEU-2 \textbf{(B-2)}, ROUGE-1 \textbf{(R-1)} and ROUGE-2 \textbf{(R-2)} to measure the similarity between generated response and the ground-truth in token and phrase level. For the overall dialogue measurement, we use a state-of-the-art well-trained model QuantiDCE~\cite{ye2021towards} in Automatic dialogue Coherence Evaluation task to measure the generated dialogue (for short, \textbf{QDCE}). In addition, we did a human evaluation to judge the knowledge appropriateness and the coherence of generated dialogues, and details are shown in~\ref{subsec:human_eval}.

\subsection{Baselines}
We compare our CET2 with the following existing approaches for knowledge-grounded conversation.
\begin{itemize}
    \item \textbf{TMN}~\cite{dinan2019wizard}. This transformer with a memory network was the baseline model, along with the release of the WoW. 
    \item \textbf{PostKS}~\cite{lian2019learning} optimizes knowledge selection performance by the knowledge prior distribution condition only on dialogue context and the posterior one given both dialogue context and gold response.
    \item \textbf{SLKS}~\cite{kim2020sequential} sequentially models knowledge selection with variables and optimize similar to \textbf{PostKS}.
    \item \textbf{PIPM}~\cite{chen2020bridging} improves SLKS by learning complement posterior knowledge information which is missing in the inference stage for SLKS. 
    \item \textbf{DukeNet}~\cite{meng2020dukenet} models knowledge tracking and knowledge shifting as dual learning. 
    \item \textbf{CoLV}~\cite{zhan2021colv} uses a collaborative latent variable model to integrate knowledge selection and knowledge-aware response generation. 
    \item \textbf{KnowledGPT}~\cite{zhao2020knowledge} compatibly combines pre-trained language models for knowledge selection and response generation.
\end{itemize}

For fairer comparison, we replaced some baselines with the same (more powerful) response generator GPT-2 as ours, such as SLKS~\cite{kim2020sequential} and DiffKS~\cite{zheng2020difference}.  This resulted in another two baselines \textbf{SLKS+GPT2} and \noindent\textbf{DiffKS+GPT2}. 

Another recent work DIALKI~\cite{wu2021dialki} regarded knowledge selection as knowledge identification in a long document, which exploits extra knowledge position information in wiki articles. We omit it here as it is unfair to compare this method with previous work such as SLKS and CoLV. In fact, by adding the extra position information, our method also outperformed DIALKI, reaching 34.5/35.6 in terms of ACC on Test Seen and Unseen of WoW.

\subsection{Implementation Details}

Our implementation was based on Pytorch~\cite{paszke2019pytorch}. We run our experiments on 2 Geforce RTX 3090 GPUs. 
For the implementation of BERT(110M) and GPT-2(117M), we used the package from the Huggingface Transformers\footnote{https://github.com/huggingface/transformers}~\cite{wolf-etal-2020-transformers}.
The training batch size and initial learning rate for BERT and GPT-2 were 4 and 32, 1e-5 and 5e-5, respectively. In the knowledge selector, the learning rate for modules other than BERT is 1e-4. 
Each knowledge is concatenated with dialogue context as the input to the BERT model with a max length of 96. The hidden size of the attention aggregation module for context embedding is the same as its input size, which is also the same as the output size of BERT. 
The input size of GAT is 768, and the headers of GAT are 8, so we set the hidden size to 96 to keep the output size of GAT the same as the input. 
A position-wise feed-forward layer~\cite{vaswani2017attention} is adopted after the GAT with the hidden size of 2048. 
The dropout rates for both GAT and feed-forward layers are 0.5 and 0.1, respectively. 
The hidden size for the pointer-based reasoning module is the same as the input size, 768 for Bert embedding. 
In training, the temperature of Gumbel-softmax is fixed to 1. 
It took around 5 and 10 epochs to achieve the reported performance for knowledge selection and response generation with optimizer Adam~\cite{kingma2015adam}. We set the balance coefficients $\lambda$ and $\mu$ for $L_{sc} $ and $ L_{g}$ to 0.5 and 2. 

For response generation, we gradually reduced the ratio of ground-truth knowledge in training following~\cite{zhao2020knowledge}. Specifically, the ratio $r$ of using ground-truth in generation will decay with training step $s$ in the rate $\beta$, which can be formulated as $r= 1/e^{s\beta}$. We set $\lambda=1e-5$ following~\cite{zhao2020knowledge}. This process facilitates generation with the right knowledge provided and has been proven to generate better responses.
As for other baselines, all the results are based on the code they provided. All the source codes and hyper-parameters settings will be released for reproduction.

\subsection{Discussion}
In this section, we analyze our experiments from five research questions.
\begin{table*}[t]
  % \setlength{\belowcaptionskip}{-0.4cm}
  % \renewcommand\tabcolsep{2.1pt}
%   \setlength\tabcolsep{3.5pt}
%   \small
  \centering \small
  \caption{Experimental results on WoW~\cite{dinan2019wizard} dataset. ACC is knowledge selection accuracy. B-1/2 are BLEU-1/2. R-1/2 refers to ROUGE1/2 scores. The methods in the upper 3 lines didn't use BERT as the encoder. Some cells with "-" indicate no available code for re-implementing the results.}
  % \resizebox{0.6\linewidth}{!}{
  \begin{tabular}{lcccccccc}
    \cmidrule[\heavyrulewidth]{1-9}
    % & \multicolumn{6}{c|}{\textbf{Selection}} & \multicolumn{8}{c}{\textbf{Response Generation}} \\
    & \multicolumn{4}{c}{\textbf{Seen}} & \multicolumn{4}{c}{\textbf{Unseen}} \\
    \cmidrule(lr){2-5}\cmidrule(lr){6-9}
    \textbf{Model} & \textbf{ACC} & \textbf{uF1} & \textbf{B-1/B-2}  & \textbf{R-1/R-2} & \textbf{ACC} & \textbf{uF1} & \textbf{B-1/B-2} & \textbf{R-1/R-2}\\
    \midrule
    TMN & 23.2  & 17.7 & 16.76.7 & - & 12.2 & 14.4 & 14.7/4.8 & -  \\
    PostKS & 23.4 & 18.1 & 17.2/7.0 & - & 9.4 & 13.5 & 15.9/6.1 & - \\
    DiffKS+GPT2 & 25.6 & 21.1 & 18.8/10.2 & 23.9/7.9  & 18.6 & 18.6 & 17.4/8.6 & 21.3/5.8 \\
    \cmidrule{1-9}
    % BERT+PoKS & 25.5  & 17.8 & - & -& 14.1 & 13.4 & -& - \\    
    % KIC & -  &18.9 & 17.3/10.5 & - & 17.3 & 16.5/9.5 \\
    PIPM & 27.8 & 19.9 & - & 19.3/7.4 & 19.4 & 17.6 & - & 17.6/5.5  \\
    DukeNet & 26.4 & 19.3 & 18.0/7.50 &\textbf{25.2}/6.8 & 19.6 & 17.1 & 16.3/6.0 & 23.3/5.3 \\
    SLKS & 26.8  & 19.3 & 18.9/10.9 & 21.1/7.0 & 18.3 & 16.1 & 17.3/8.0 &  18.2/5.9 \\
    SLKS+GPT2 & 26.8 & 20.5 & 18.8/9.9 &23.4/7.4 & 18.3 & 17.7 & 16.4/7.7  & 20.3/5.2 \\
    KnowledGPT & 28.0  & 21.9 & 19.5/10.8 & 24.7/8.5 & 25.4 & 19.6 & 17.7/9.1 &  22.3/6.5 \\
    CoLV & 30.1 & - & - &20.6/7.9  & 18.9 & - & - & 19.7/6.3\\
    \cmidrule{1-9}
    CET2 & \textbf{31.7}  & \textbf{22.4} & \textbf{20.2}/\textbf{11.4}  & \textbf{25.2}/\textbf{8.9} & \textbf{30.1} & \textbf{21.1} & \textbf{19.2}/\textbf{10.4} & \textbf{23.7}/\textbf{7.6} \\
    \cmidrule[\heavyrulewidth]{1-9}
  \end{tabular}
  % }
  
  \label{table:wow}
 \end{table*}
 
 \begin{table}[t]
  \setlength{\belowcaptionskip}{-0.4cm}
  \renewcommand\tabcolsep{2.7pt}
  \centering
  \caption{Knowledge transition and dialogue coherence evaluations on WoW.\textit{Adh.} and \textit{Div.} are the metrics to evaluate the knowledge accuracy of selecting adhesive and diverse knowledge, respectively.
  QDCE~\cite{ye-etal-2021-towards-quantifiable} measures coherence between dialogue context and the generated response.}
  \resizebox{0.9\linewidth}{!}{
  \begin{tabular}{lcccccc}
    \cmidrule[\heavyrulewidth]{1-7} & \multicolumn{3}{c}{\textbf{Seen}} & \multicolumn{3}{c}{\textbf{Unseen}} \\
    \cmidrule(lr){2-4}\cmidrule(lr){5-7}
    \textbf{Model} & \textbf{Div.} & \textbf{Adh.} & \textbf{QDCE} & \textbf{Div.} & \textbf{Adh.} & \textbf{QDCE}\\
    \cmidrule{1-7}
    DukeNet & 16.4 & 51.3 & 3.205 & 10.6 & 45.8 & 3.215 \\
    SLKS & 13.2 & 54.5 & 3.298 & 6.9 & 39.9 & 3.294   \\
    DiffKS & 15.2 & 49.7 & 3.211 & 11.1 & 36.0 & 3.204  \\
    KnowledGPT & 16.5 & 49.8 & 3.351 & 11.5 & 56.5 & 3.363 \\
    \cmidrule{1-7}
    CET2 & \textbf{20.4} & \textbf{57.9} & \textbf{3.445} & \textbf{18.0} & \textbf{57.7} & \textbf{3.459} \\
    \cmidrule[\heavyrulewidth]{1-7}
  \end{tabular}
  }
  \label{table:transition}
 \end{table}
 
\noindent\textbf{Q1. Is CET2 able to perform well on knowledge selection and response generation?}
% \subsubsection{Q1. Is CET2 able to perform well on knowledge selection and response generation}

The experimental evaluation results on WoW and Holl-E are shown in Table \ref{table:wow}, \ref{table:transition} and \ref{table:holle}. In Table \ref{table:wow}, \ref{table:transition}, our CET2 model achieves best results compared to all the baselines in terms of all metrics on the Seen and Unseen test sets of WoW.

For knowledge selection, CET2 outperforms the very recent work CoLV in ACC by margins of $1.6$ and $11.2$ on Test Seen and Test Unseen, respectively. Even compared with KnowledGPT, which has the best comprehensive performance on both test sets, our method also improved by $3.7$ and $4.7$, reaching 31.7/30.1. It is particularly worth noting that our method exceeds five strong baselines SLKS, DiffKS, DukeNet, PIPM, and CoLV on the more difficult WoW Test Unseen dataset by $11.8$, $11.5$, $10.5$, $10.7$, and $11.2$, respectively, presenting powerful 
generalization ability owing to our appropriate topic transition design.

For response generation, CET2 also achieves the best results on all automatic metrics. Moreover, compared with SLKS+GPT2, DiffKS+GPT2, and KnowledGPT using the same response generator GPT2, our model still improves generation significantly, demonstrating that higher knowledge selection accuracy does lead to better generation performance. Note that KnowledGPT also uses the same encoder BERT and decoder GPT2 as CET2. CET2 still outperforms KnowledGPT even though the latter adopts more sophisticated and costly training strategies (the reinforcement step and the curriculum step). 

Similar results can also be observed on Holl-E in Table \ref{table:holle}. CET2 obtains significant performance gains on all the metrics compared to other baselines, showing the highest accuracy in knowledge selection with margins of $8.5$, $4.2$, $7.7$, $7$, and $5$ with respect to five strong baselines SLKS, DiffKS, DukeNet, PIPM, and CoLV. 

\begin{table}[t]
  \setlength{\belowcaptionskip}{-0.4cm}
  \renewcommand\tabcolsep{3pt}
  \centering
  \caption{Evaluation results on Holl-E dataset. The best results are highlighted with \textbf{bold}.B-1/B-2 and R-1/R-2 are BLEU1/2 and ROUGE1/2.}
  \resizebox{0.8\columnwidth}{!}{
  \begin{tabular}{lcccc}
  \cmidrule[\heavyrulewidth]{1-5}
    % [3pt]
    \textbf{Model} & \textbf{ACC} & \textbf{uF1} & \textbf{B-1/B-2}  & \textbf{R-1/R-2}\\
    \cmidrule{1-5}
    % TMN & 22.7 & 20.1 & - & - \\
    % PostKS & 22.5 & 19.9 & - & - \\
    % BERT+PostKS & 27.6 & 27.8 & - & - \\
    SLKS & 29.2 & 29.8 & 28.0/22.2 & 31.3/23.2 \\
    % SLKS+GPT2~\cite{zhao2020knowledge} & 29.2 & - & - & - \\
    DiffKS+GPT2 & 33.5 &  31.9 & 31.2/26.9 & 33.9/24.7   \\
    PIPM & 30.7 & 30.8 & - & - \\
    DukeNet & 30.0 & 30.6 & 30.1/22.5 & \textbf{36.5}/23.0 \\
    CoLV & 32.7 & - & - & 32.0/\textbf{25.8} \\
    CET2 & \textbf{37.7} & \textbf{32.9} & \textbf{31.8}/\textbf{28.0} & 34.8/25.6 \\
  \cmidrule[\heavyrulewidth]{1-5}
  \end{tabular}
  }

  \label{table:holle}
 \end{table}
  
\noindent\textbf{Q2. Whether CET2 improves topic coherence and knowledge diversity?} 

As shown in Table \ref{table:transition}, compared with four strong baselines, CET2 obtains the highest score on Div. and Adh. , with margins of $4\%/7.4\%$, $7.2\%/11.1\%$, $5.2\%/6.9\%$, and $3.9\%/6.5\%$ on two test sets of WoW, which proves that CET2 indeed improves both dialogue topic adhesion and diversity during dialogue knowledge flow modeling. Moreover, the result of QDCE, which measures the overall dialogue coherence from the generated conversations, further validates that our CET2 is able to capture better dialogue transition logic.

\noindent\textbf{Q3. Whether each module contributes to CET2 Performance?}

We conduct a series of ablation experiments on the WoW dataset. Four variants are designed for ablation study as follows:
\begin{itemize}
    \item \textit{w/o ShiftLoss}: removing the Topic Shifting Constraint Loss;
    \item \textit{w/o CrossOpt}: cutting the topic diversity feature calculated by the cross operation between knowledge candidates and previously selected knowledge;
    \item \textit{w/o CoherOpt} removing topic diversity feature from coherence operator;\
    \item \textit{w/o PointerNet}: replacing Comparative Knowledge Selection with a simpler knowledge selection module, where knowledge scores are the attention scores between the context representation $\mathbf{c}$ and each transition-aware knowledge representation from $\mathbf{E}$.
\end{itemize}    
Almost all the results of these variants, as shown in Table \ref{table:wow_abla}, exhibit performance drops in knowledge selection and response generation, validating their contributions to our model's generalization ability.

\noindent\textbf{Q4. Which parts of CET2 improve coherence and diversity?} 

In Table~\ref{table:wow_abla}, the metric \textbf{Div.} becomes worse no matter which module is removed in two testing sets. For \textbf{Adh.}, it drops for all variants in the seen test set while increases for variants \textit{w/o shiftLoss} and \textit{w/o CrossOpt} in the Unseen test set, which demonstrates that shift loss and cross operator somehow advocate diversity and suppress adhesion in cases where the model needs stronger generalization ability.  Besides, if we look at the experiment \textit{w/o CoherOpt}, \textbf{Adh.} drops the most compared to other variants, showing that the coherence operator is mainly in charge of adhesion.

\noindent\textbf{Q5. How does CET2 improve the knowledge selection?}

\begin{figure*}[t]
\centering
  \includegraphics[width=0.95\linewidth]{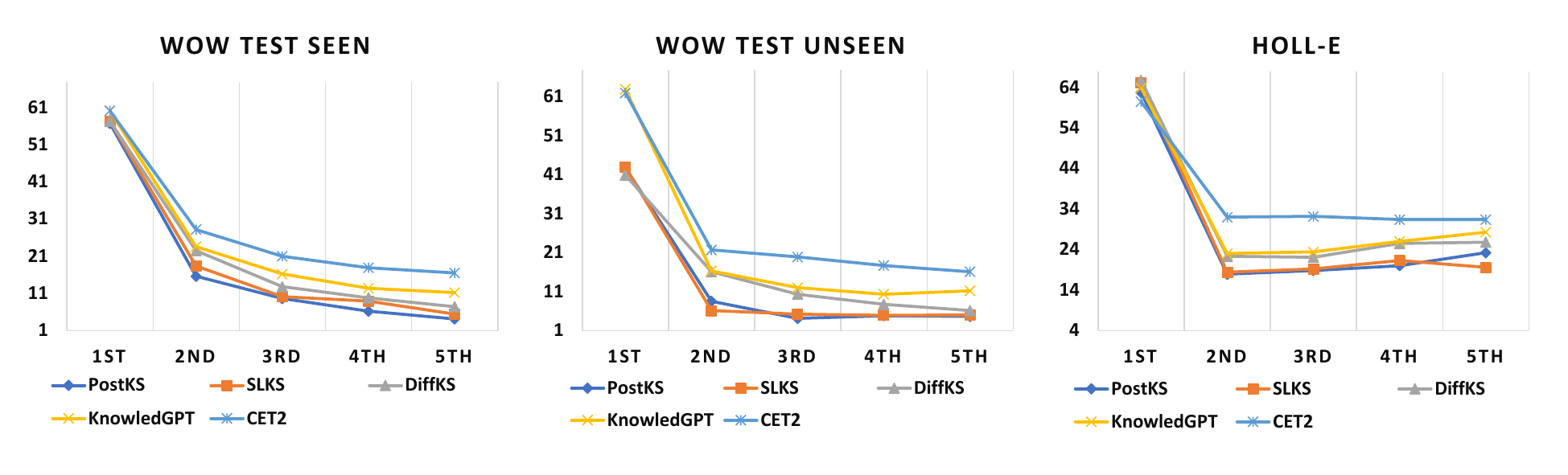}
\caption{Knowledge Selection accuracy overturns. The horizontal axis is the turn index, and the vertical axis is knowledge selection accuracy. The knowledge selection accuracy drops as conversations progress longer.}
\label{fig:turn_acc}
\end{figure*}

In Table~\ref{table:wow}, CET2 displays obvious superiority on unseen generalization. We are curious about what CET2 improves about knowledge selection. Thus we investigate the fine-grained knowledge selection performance scattering different turns. Specifically, we divide data samples into five turn groups, with the i-th group containing only data samples from the i-th turn of conversations. Within each group, we compute the knowledge selection accuracy, and the results are shown in Figure~\ref{fig:turn_acc}. From these line charts, we find that CET2 performs similarly with other baselines for first-turn samples while presenting superior capability in the subsequent turns. In fact, knowledge selection can be more challenging as the conversation goes on. The difficulty mainly comes from the more complex dialogue context and the higher possibility of various topic transitions.

\begin{table}[t]
  \setlength{\belowcaptionskip}{-0.4cm}
  \small
  \centering  
  \caption{Ablation test results on WoW dataset. Almost all parts contribute to the CET2 final performance in the four metrics. One exception is on Adh. Adh. increases after removing Shift loss or Cross Operator.}
  \footnotesize
    \renewcommand\tabcolsep{2.2pt}
  \resizebox{\columnwidth}{!}{
  \begin{tabular}{lcccccccc}
  \cmidrule[\heavyrulewidth]{1-9}
    & \multicolumn{4}{c}{\textbf{Seen}} & \multicolumn{4}{c}{\textbf{Unseen}}  \\
    % \cmidrule{2-9}
    \cmidrule(lr){2-5}\cmidrule(lr){6-9}
    \textbf{Model} & \textbf{ACC} & \textbf{Div.} & \textbf{Adh.} &\textbf{uF1} & \textbf{ACC} & \textbf{Div.} & \textbf{Adh.} & \textbf{uF1} \\
    \cmidrule{1-9}
    \textbf{CET2} & 31.7 & 20.4 & 57.9 & 22.4 & 30.1 & 18.0 & 57.7 & 21.1 \\
    % \cmidrule{1-9}
    \quad +w/o ShiftLoss & 30.6 & 19.5 & 56.0 & 22.0 & 29.5 & 17.0 & 58.6 & 21.3 \\
    \quad +w/o CrossOpt & 30.2 & 18.9 & 56.4 & 21.9 & 28.4 & 14.8 & 59.0 & 20.6 \\
    \quad +w/o CoherOpt &  30.0 & 19.3 & 54.8 & 21.9 & 28.8 & 16.6 & 56.8 & 20.8\\
    \quad +w/o PointerNet & 29.7 & 18.6 & 55.5 & 21.8 & 29.6 & 17.2 & 57.7 & 21.1 \\
  \cmidrule[\heavyrulewidth]{1-9}
  \end{tabular}
  }
 \label{table:wow_abla}
\end{table}

\subsection{Human Evaluation}
\label{subsec:human_eval}
We conduct human evaluations to complement automatic evaluations. We published the evaluation task on Amazon Mechanical Turk and required only native speakers to evaluate generated responses. Precisely, each evaluation sample consists of the dialogue context, four responses from ours, KnowledGPT, SLKS, and DukeNet evaluated by three different annotators. First, these curators were given instructions for criterion definitions and evaluation steps. Then they will go through the dialogue context, compare each pair of responses, and choose a better one from naturalness and appropriateness.

The human evaluation criteria consisted of \textbf{naturalness} and \textbf{appropriateness}. The former emphasizes the readability and fluency of the generated response, while the latter highlights whether appropriate knowledge is used in the response given the conversation context, which emphasizes both knowledge coherence and engagement. We randomly select 300 samples from the seen and unseen test set of WoW, and three curators evaluate each sample on Amazon Mechanical Turk. The results are shown in Table~\ref{table:wow_humaneval}. Our method significantly outperforms SLKS and DukeNet in both criteria. For knowledGPT, although CET2 and knowledGPT both use GPT-2 as the generator, CET2 still performs better. We compute the FLeiss' Kappa~\cite{fleiss1971measuring} to measure the agreement of all the curators on each sample. As more than 20 people participated in this evaluation study, we have a moderate fleiss' kappa value within the range 0.3-0.5.

\begin{table}[t]
  \setlength{\belowcaptionskip}{-0.4cm}
  \centering   \small
   \caption{Human evaluation on WoW dataset. \emph{Win} and \emph{Lose} is the percentage of CET2 wins or loses compared to other methods. The remaining part of \emph{tie} is omitted.}
    \renewcommand\tabcolsep{4pt}
  \resizebox{0.95\columnwidth}{!}{
  \begin{tabular}{lcccccc}
  \cmidrule[\heavyrulewidth]{1-7}
    & \multicolumn{3}{c}{\textbf{Naturalness}} & \multicolumn{3}{c}{\textbf{Appropriateness}}  \\
    \cmidrule(lr){2-4}\cmidrule(lr){5-7}
    \textbf{Methods} & \textbf{Win} & \textbf{Lose} & \textbf{$\kappa$} & \textbf{Win} & \textbf{Lose} & \textbf{$\kappa$} \\
    \cmidrule{1-7}
    \multicolumn{7}{l}{\textbf{WoW (Seen)}} \\
    \cmidrule{1-7}
    CET2 vs. SLKS  & 85 & 7 & 0.43 & 70 & 10 & 0.43 \\
    CET2 vs. KnowledGPT  & 34 & 18 & 0.40 & 30 & 12 & 0.38 \\
    CET2 vs. DukeNet  & 79 & 11 & 0.50 & 74 & 11 & 0.43 \\
    \cmidrule{1-7}
    \multicolumn{7}{l}{\textbf{WoW (Unseen)}} \\
    \cmidrule{1-7}
    CET2 vs. SLKS  & 78 & 6 & 0.32 & 65 & 12 & 0.31 \\
    CET2 vs. KnowledGPT  & 28 & 17 & 0.31 & 24 & 18 & 0.32 \\
    CET2 vs. DukeNet  & 69 & 11 & 0.45 & 61 & 11 & 0.32 \\
  \cmidrule[\heavyrulewidth]{1-7}
  \end{tabular}

  }
 \label{table:wow_humaneval}
\end{table}
% \vspace{-15pt}

\section{Case Study}
\begin{figure}[t]
\centering
  \includegraphics[width=\linewidth]{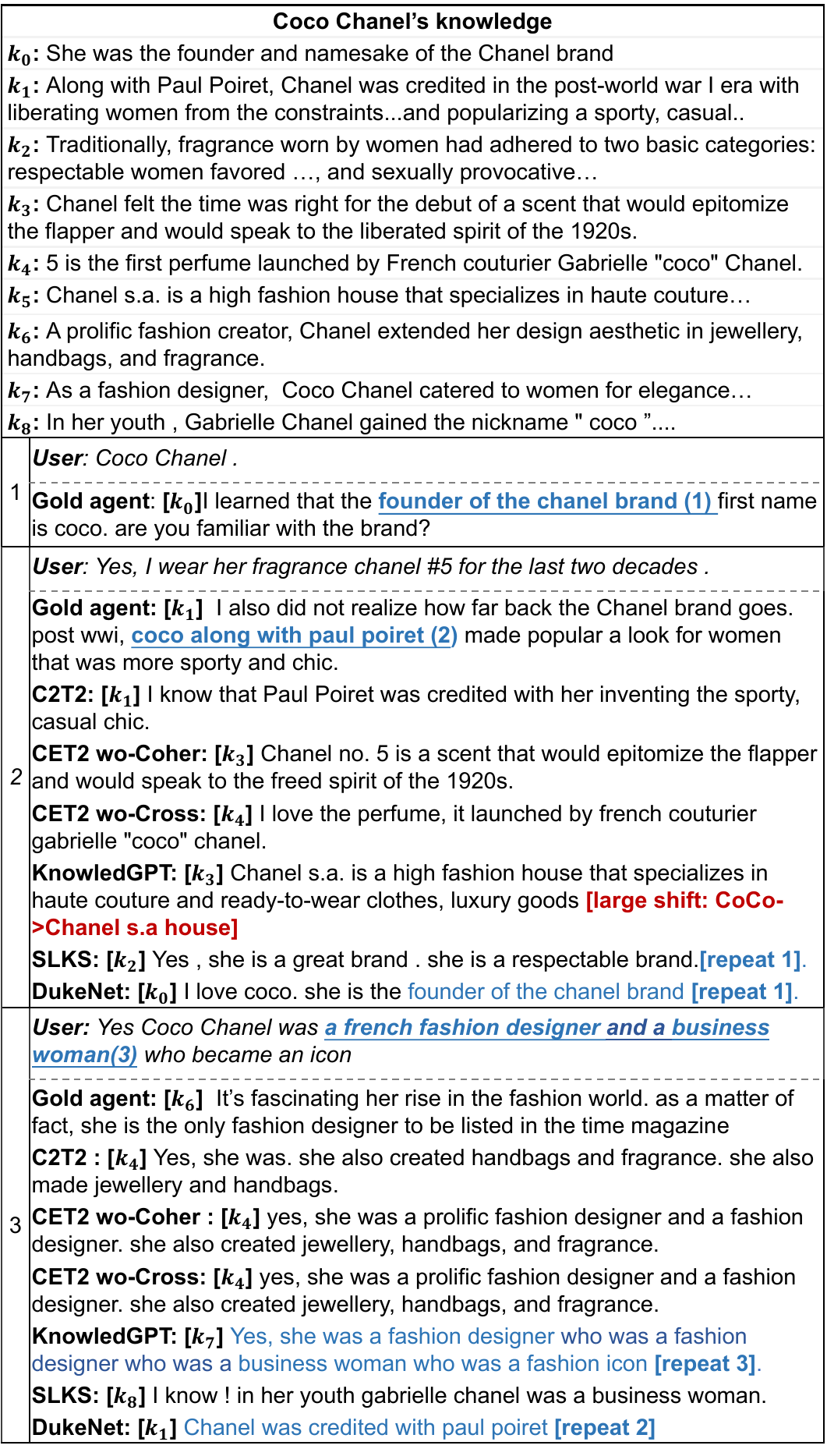}
\caption{Generations of different methods, "[$k_i$]" indicates the chosen knowledge. Blue words indicate repetition and red words are related to incoherence. The numbers following blue words point to their repeated parts with the same numbers in a dialogue history. \textbf{\textit{Post i}} is the dialogue history at turn \textit{i}. At each turn \textit{i}, models try to predict the response of \textbf{\textit{GOLD}}. }
\label{fig:case_study_coco}
\end{figure}

To give a more intuitive evaluation of the generated conversations, we visualized a complete conversation about the topic \emph{Coco Chanel}, as shown in Figure~\ref{fig:case_study_coco}. In this figure, $k_1$ to $k_9$ are part of the list of knowledge candidates, and there are three turns of dialogue in total. Sentences starting with \textbf{\textit{Post}} are utterances from the user. The lines below post are the gold response and generated responses of different methods, including CET2, KnowledGPT, SLKS, and DukeNet. To show the effects of CET2's different modules, we also give the results of ablated CET2, CET2 wo-Coher, and CET2 wo-Cross.   \emph{GOLD} indicates the expected gold response from the wizard user(the agent we need to model in the KGC task), who tends to convey more knowledge to the apprentice. The indicators in the brackets ``[]'' leading these responses, e.g. ``[$k_4$]", are the knowledge sentences selected by these methods.
\begin{itemize}
    \item For the first turn, all the methods chose the right knowledge and generated reasonable results, so we just ignore the comparisons here and only give the gold responses as the reference for the following turns.
    \item For the \emph{Post 2}, only our CET2 predicts the right knowledge "[$k_1$]". SLKS and DukeNet also provide relevant responses but with limited information since both methods just repeat the content in the dialogue history,  ``founder of the channel brand(1)", which deviates from the target of knowledge-grounded conversations to convey new knowledge. 
    KnowledGPT also introduces new information, but the knowledge shift is too large from the topic \emph{CoCo Chanel} fragrance to the topic of \emph{Chanel s.a house}, leading to an unhuman conversation and possibly disengaging user. As for our ablated experiments, we found CET2 wo-Coher introduced ``Chanel \#5" from a totally new perspective ``the scent", while wo-Cross tended to continue the conversation with more coherent knowledge. This in fact aligns with our design of coherence and cross operators.
    \item For \emph{Post 3}, still only CET2 predicted the right knowledge as well as CET2 wo-Coher and wo-Cross. SLKS chose wrong knowledge and also generated less informative responses. DukeNet and Knowledge just repeat the agent's utterance in history.
\end{itemize}

\section{Conclusion}
Effective topic transition modeling is critical for a coherent and engaging conversation. In this paper, we proposed a new method named CET2 that simultaneously considered topic coherence and knowledge diversity for topic transitions in knowledge-grounded conversations. CET2 selects appropriate knowledge by using context-aware representations and devising transition-aware features of the knowledge candidates, effectively performing a comparative knowledge selection module, and adopting a variance-aware training strategy. Our CET2 method achieved new state-of-the-art performance on both Wizard of Wikipedia and Holl-E, in particular, made significant progress in the knowledge selection task, especially in the harder unseen case. However, measuring knowledge coherence and the appropriateness of diversity is a non-trivial problem. In this paper we simply use accuracy of the knowledge change and knowledge adhesion to reflect these two properties, which could be replaced by more well-defined metrics.

\section*{Acknowledgments}
We thank all reviewers for their valuable comments. This research/project is supported by the National Research Foundation, Singapore under its Industry Alignment Fund – Pre-positioning (IAF-PP) Funding Initiative and AI Singapore Programme (AISG Award No: AISG-GC-2019-001-2A). Any opinions, findings and conclusions or recommendations expressed in this material are those of the author(s) and do not reflect the views of National Research Foundation, Singapore.
21\cite{devlin-etal-2019-bert}
25
33\cite{}

% \clearpage
\bibliography{custom, anthology}
\bibliographystyle{IEEEtran}

\begin{IEEEbiography}[{\includegraphics[width=1in,height=1.25in,clip,keepaspectratio]{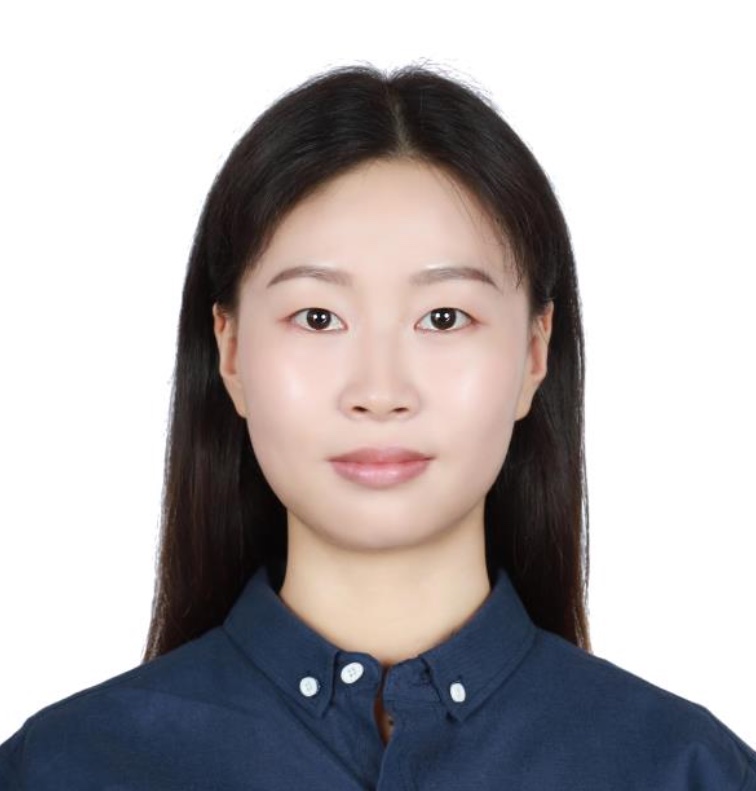}}]{Lin Xu} received the Bachelor's degree in Software Engineering and Master's degree in Computer Science from Wuhan University and Sun Yat-Sen University in 2017 and 2020, respectively. She is currently a PhD student at National University of Singapore since 2020. Her research interests include specialized knowledge-grounded dialogue systems, interactions among multiple intelligent agents based on large language models, and multimodal-to-language generations.
\end{IEEEbiography}
\begin{IEEEbiographynophoto}{Qixian Zhou} got a bachelor's degree in software engineering and a master's degree in computer science in 2017 and 2019, respectively, from Sun Yat-sen University.
He used to work at DMAI, a company founded by Professor Song-Chun Zhu from UCLA, in 2019, responsible for the implementation of various AI projects. 
He joined ByteDance as an algorithm engineer in 2022, focusing on NLP algorithm research and development in the gaming field. His main research direction is generative AI, large language models, multimodal recognition, and other fields.
\end{IEEEbiographynophoto}
\begin{IEEEbiography}[{\includegraphics[width=1in,height=1.25in,clip,keepaspectratio]{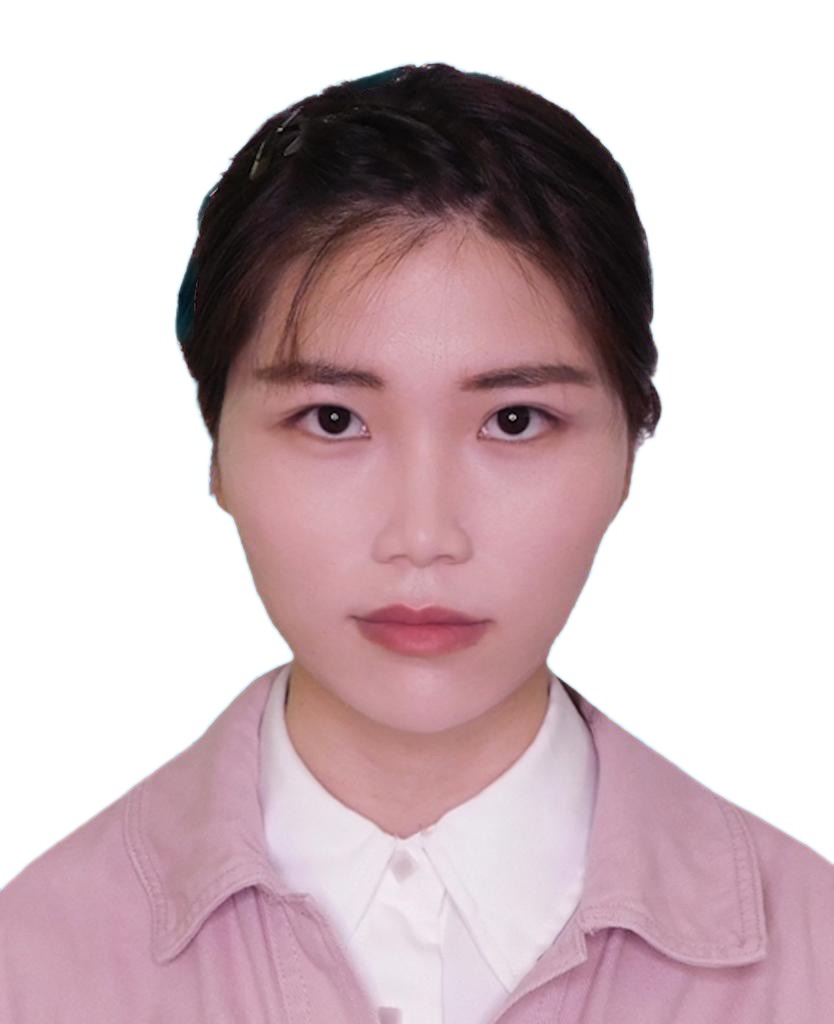}}]{Jinlan Fu} earned her PhD in computer science from Fudan University. She is currently a research fellow at the Institute of Data Science, National University of Singapore. Her research interests include natural language processing, text generation and evaluation, as well as interpretable analysis.
\end{IEEEbiography}
\begin{IEEEbiography}[{\includegraphics[width=1in,height=1.25in,clip,keepaspectratio]{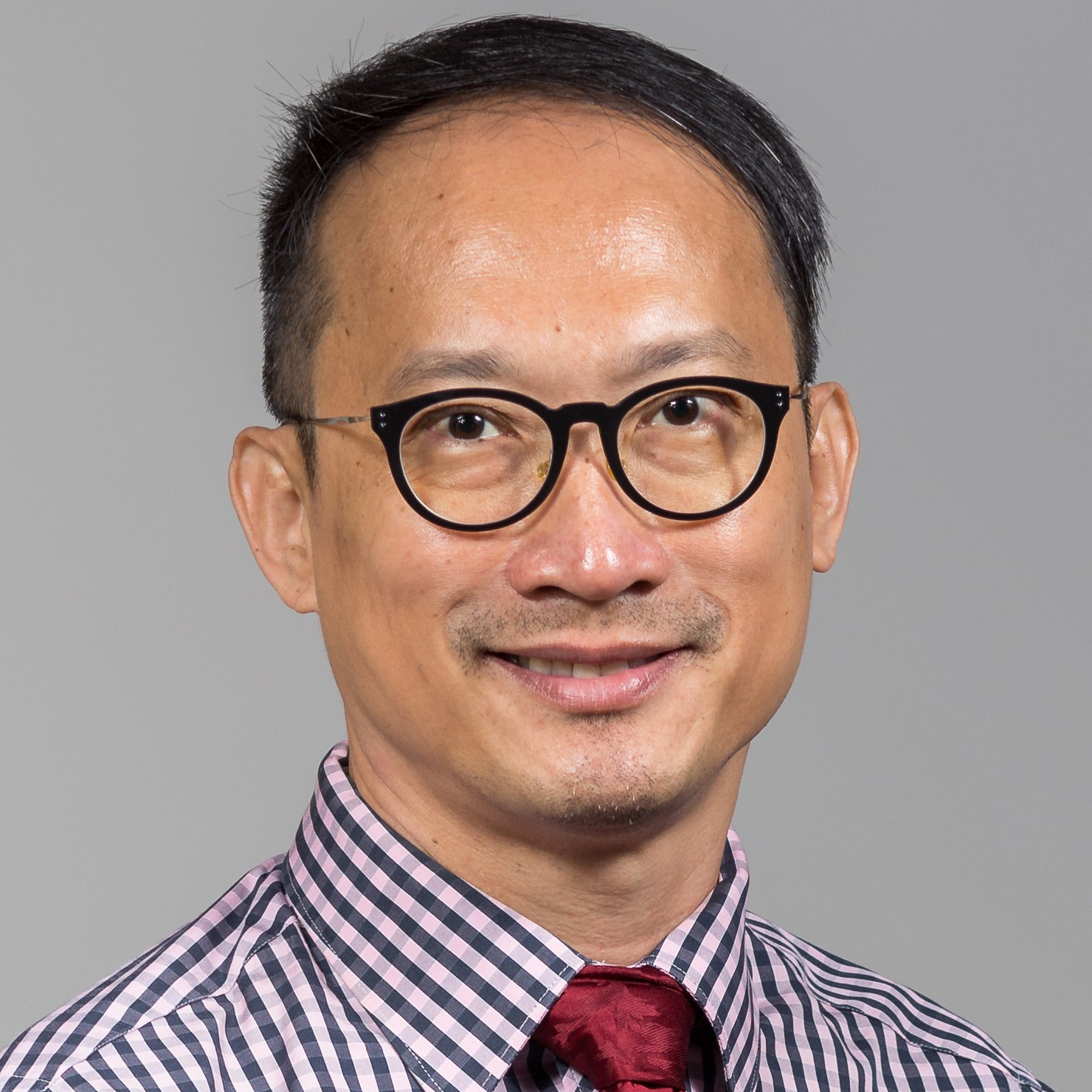}}]{See Kiong Ng} received his B.S. degree in Computer Science from Carnegie Mellon University (CMU), M.S. degree from University of Pennsylvania, and Ph.D. degree in Computer Science from CMU.
He is currently a Professor of Practice with the School of Computing, National University of Singapore, where he is also the Director of translational research with
the Institute of Data Science. He has authored over 130 papers in diverse and cross-disciplinary research topics from bioinformatics to smart cities based on data science and AI.
\end{IEEEbiography}

\vfill

\end{document}